\documentclass[10pt,twocolumn,twoside]{IEEEtran}
\usepackage[utf8]{inputenc}         
\usepackage[english]{babel}
\usepackage{multirow}
\usepackage{booktabs}
\usepackage{mwe}
\usepackage{tabularx}
\usepackage[normalem]{ulem}

\usepackage{cite}

\usepackage[
    type={CC},
    modifier={by-nc-sa},
    version={3.0},
]{doclicense}

\ifCLASSINFOpdf
\usepackage{graphicx}
\usepackage{adjustbox}
\else
\fi
\usepackage[absolute,showboxes]{textpos}
\TPMargin{5pt}
\newcommand{\copyrightstatement}{
    \begin{textblock}{0.8}(0.08,0.02)    
         \noindent
         \footnotesize
         Accepted manuscript for Biomedical Signal Processing and Control. DOI: https://doi.org/10.1016/j.bspc.2020.102037 \\ 
         \copyright 2020. This manuscript version is made available under the CC-BY-NC-ND 4.0 license http://creativecommons.org/licenses/by-nc-nd/4.0/
    \end{textblock}
}

\usepackage{subcaption}

\usepackage{url}

\usepackage{dblfloatfix}
\usepackage{multirow}
\usepackage{balance}
\usepackage{cite}
\usepackage{capt-of}
\usepackage[cmex10]{amsmath}
\usepackage{amssymb}

\usepackage{tabularx}
\usepackage{textcomp}
\usepackage{bm}
\usepackage{calc}

\usepackage{xcolor,colortbl}

\usepackage{tabu}
\usepackage{mdframed}
\usepackage{makecell}
\makeatletter
\def\blfootnote{\xdef\@thefnmark{}\@footnotetext}
\makeatother
\setcellgapes{1.5pt}
\newcolumntype{Y}{>{\centering\arraybackslash}X}

\usepackage[ruled,lined]{algorithm2e}

\begin{document}
	\bstctlcite{IEEEexample:BSTcontrol}
	\title{Intra- and Inter-epoch Temporal Context Network (IITNet) Using Sub-epoch Features for Automatic Sleep Scoring on Raw Single-channel EEG}
	%
	%
	%

	\author{
        Hogeon Seo$^*$,
        Seunghyeok Back$^*$, 
        Seongju Lee$^*$,
        Deokhwan Park,
        Tae Kim,
        Kyoobin Lee†
        \thanks{\text{*}  These authors contributed equally to the paper.}
		\thanks{H. Seo is with Korea Atomic Energy Research Institute, 111, Daedeok-daero 989beon-gil, Yuseong-gu, Daejeon 34057, Republic of Korea.\newline
		S. Back, S. Lee, D. Park, and K. Lee are with the School of Integrated Technology (SIT), Gwangju Institute of Science and Technology (GIST), Cheomdan-gwagiro 123, Buk-gu, Gwangju 61005, Republic of Korea. \newline
		T. Kim is with the Department of Biomedical Science and Engineering, and School of Life Sciences, Gwangju Institute of Science and Technology (GIST), Cheomdan-gwagiro 123, Buk-gu, Gwangju 61005, Republic of Korea.}
		\thanks{†Corresponding author: Kyoobin Lee {\tt\footnotesize kyoobinlee@gist.ac.kr}}}

	\maketitle
	\copyrightstatement
	\begin{abstract}	
     A deep learning model, named IITNet, is proposed to learn intra- and inter-epoch temporal contexts from raw single-channel EEG for automatic sleep scoring. To classify the sleep stage from half-minute EEG, called an epoch, sleep experts investigate sleep-related events and consider the transition rules between the found events. Similarly, IITNet extracts representative features at a sub-epoch level by a residual neural network and captures intra- and inter-epoch temporal contexts from the sequence of the features via bidirectional LSTM. The performance was investigated for three datasets as the sequence length (\(L\)) increased from one to ten. IITNet achieved the comparable performance with other state-of-the-art results. The best accuracy, MF1, and Cohen's kappa ($\kappa$) were $83.9\%$, $77.6\%$, $0.78$ for SleepEDF (\(L\)=10), $86.5\%$, $80.7\%$, $0.80$ for MASS (\(L\)=9), and $86.7\%$, $79.8\%$, $0.81$ for SHHS (\(L\)=10), respectively. Even though using four epochs, the performance was still comparable. Compared to using a single epoch, on average, accuracy and MF1 increased by $2.48\%p$ and $4.90\%p$ and F1 of N1, N2, and REM increased by $16.1\%p$, $1.50\%p$, and $6.42\%p$, respectively. Above four epochs, the performance improvement was not significant. The results support that considering the latest two-minute raw single-channel EEG can be a reasonable choice for sleep scoring via deep neural networks with efficiency and reliability. Furthermore, the experiments with the baselines showed that introducing intra-epoch temporal context learning with a deep residual network contributes to the improvement in the overall performance and has the positive synergy effect with the inter-epoch temporal context learning.

	\end{abstract}
	
	\begin{IEEEkeywords}
	    deep learning, classification, single-channel EEG, sleep scoring, sequence, temporal context, end-to-end. 
	\end{IEEEkeywords}

	%
	\IEEEpeerreviewmaketitle
	
	\section{Introduction}
	\label{sec:intro}

    Sleep scoring, also known as "sleep stage classification" and "sleep stage identification," is essential for diagnosis and treatment of sleep disorders \cite{wulff2010sleep}. Many individuals suffering from sleep disorders are at risk of underlying health problems \cite{torabi2015withstanding}. Typical sleep disorders (e.g., sleep apnea, narcolepsy, and sleepwalking) can be diagnosed via polysomnography (PSG) \cite{berthomier2007automatic}, the gold standard of sleep scoring. PSG is based on the biosignals of body functions such as brain activity (electroencephalogram, EEG), eye movement (electrooculogram, EOG), heart rhythm (electrocardiogram, ECG), and muscle activity of the chin, face, or limbs (electromyogram, EMG). These recorded signals are analyzed by trained human experts, who label each 20- or 30-second segment of PSG data (an "epoch") with its corresponding sleep stage. The sleep stages are classified into wakefulness (W), rapid eye movement (REM), and non-REM (NREM) following the Rechtschaffen and Kales (R\&K) rules \cite{ALLANHOBSON1969644,rechtschaffen1968manual} or the American Academy of Sleep Medicine (AASM) rules \cite{berry2012aasm}. For the AASM rules, NREM is further divided into three stages, referred to as S1, S2, and S3 or N1, N2, and N3. To draw a whole-night hypnogram showing the sleep stage as a function of sleep time, experts must visually inspect all epochs and label their sleep stages. This manual sleep scoring is labor-intensive and time-consuming \cite{stephansen2017use,stepnowsky2013scoring,rosenberg2013american}. Thus, automatic sleep scoring for healthcare and well-being  is in high demand.
    
    \begin{figure*}[ht]
        \centering
      \includegraphics[width=\textwidth]{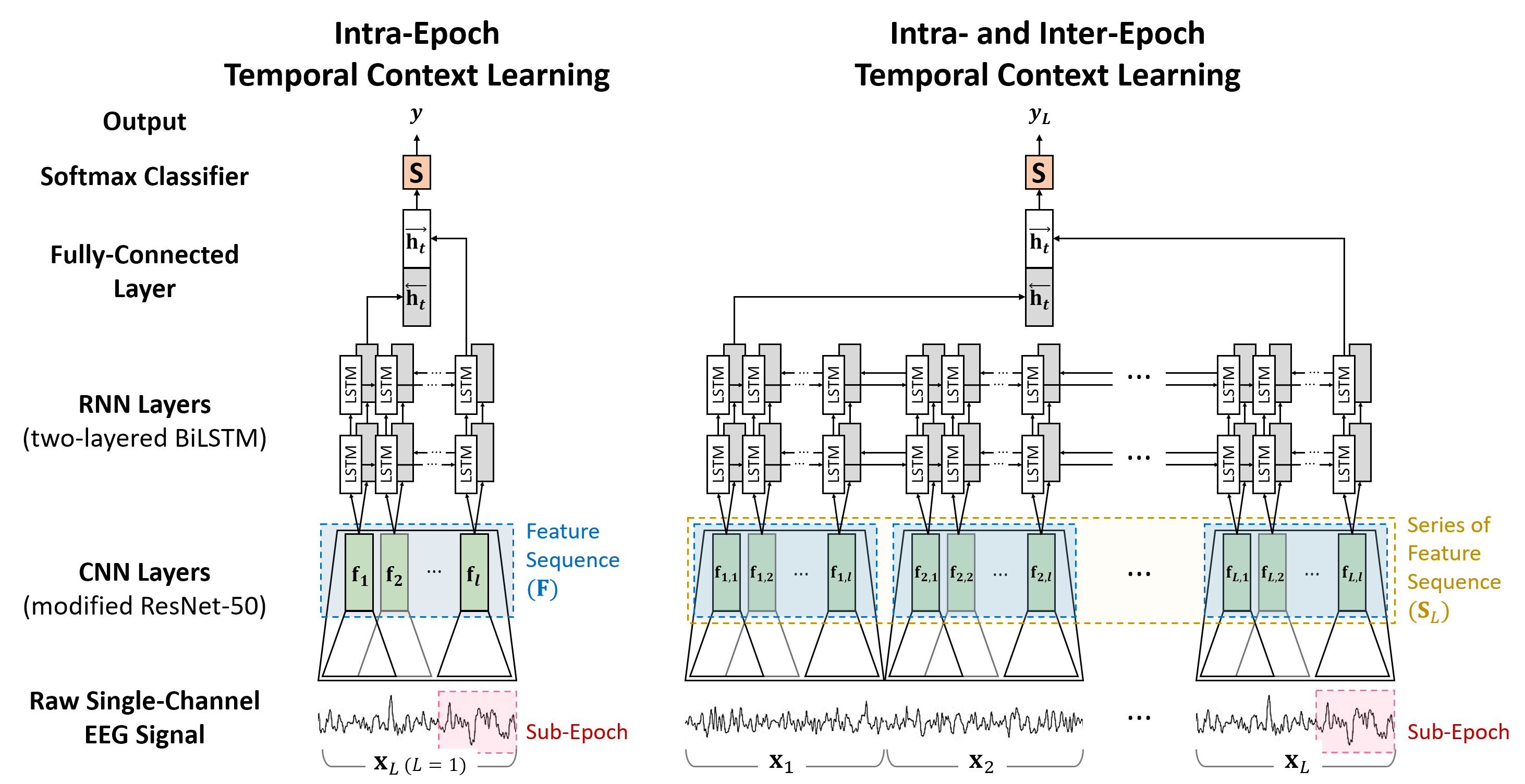}
      \caption{The architecture of the proposed model, IITNet. The left shows IITNet of one-to-one configuration for intra-epoch temporal context learning. The right shows IITNet of many-to-one configuration for intra- and inter-epoch temporal context learning. Each green box indicates the representative feature extracted from sub-epoch (red dash box). These representative features are aggregated into a feature sequence (blue dash box) in intra-epoch temporal context learning or a series of feature sequences (yellow dash box) for both the intra- and inter-epoch temporal context learning.}  
      \label{fig:ModelArchitecture}
    \end{figure*}
    
    Many sleep scoring methods for automatic analysis of PSG data have been proposed. In particular, handcrafted feature extraction techniques have been widely used \cite{aboalayon2016sleep}. The features can be extracted from time-domain signals \cite{li2017hyclasss,koley2012ensemble}, frequency/time-frequency domain data \cite{sharma2017automatic,hassan2016decision,langkvist2012sleep,fraiwan2012automated,berthomier2007automatic}, or non-linear parameters \cite{hassan2016computer,lajnef2015learning,liang2012automatic}, and have been analyzed using fuzzy classification \cite{berthomier2007automatic}, decision trees \cite{hassan2016computer}, random forest algorithms \cite{li2017hyclasss,sharma2017automatic,hassan2016decision,fraiwan2012automated}, and support vector machines \cite{lajnef2015learning,zhu2014analysis,koley2012ensemble}. In some cases, multi-channel or multi-modality data have been used \cite{stephansen2017use,lajnef2015learning}. Such previous studies have shown that the application of machine learning to handcrafted features is effective for automatic sleep scoring. However, the associated approaches may require additional handcrafted tuning to analyze PSG data obtained from different recording environments, as the features are hand-engineered based on the specific PSG system and available datasets \cite{supratak2017deepsleepnet}.
    
    Deep learning has been adopted to score sleep stages from PSG data automatically. When human sleep experts score the sleep stage of an epoch, they generally find the sleep-related event (such as a K-complex, sleep spindle, or frequency components: alpha, beta, delta, and theta activities) in the epoch. Then, they analyze the relations between the sleep-related events and sleep stages in its neighboring epochs. Thus, they inspect the data at both intra- and inter-epoch levels. In \cite{sors2018convolutional,chambon2018deep,supratak2017deepsleepnet,vilamala2017deep,tsinalis2016automatic,manor2015convolutional,wulsin2011modeling,mirowski2009classification}, features were extracted by convolutional neural networks (CNNs). Some deep neural networks have learned features from multi-channel or multi-modality data to improve sleep scoring performance \cite{chambon2018deep,o2014montreal,parra2005recipes}. Recently, recurrent neural networks (RNNs) are being adopted to consider transition rules such as those of the AASM manual and to learn temporal information from the sequence of epochs \cite{phan2019seqsleepnet,sors2018convolutional,chambon2018deep,phan2018joint,dong2018mixed,supratak2017deepsleepnet,stephansen2017use,biswal2017sleepnet,tsinalis2016automatic,hsu2013automatic}. For instance, an epoch can be labeled N2 when K-complexes or sleep spindles exist in the last half of its preceding epoch. Supratak et al. used RNNs to consider the inter-epoch temporal context between epoch-wise features individually extracted from each epoch by CNNs \cite{supratak2017deepsleepnet}. Their results showed that consideration of the transition rules by analyzing the inter-epoch temporal context is essential for automatic sleep scoring. However, this model required two-step training and just considered the inter-epoch temporal context. Phan et al. \cite{phan2019seqsleepnet} introduced RNNs to analyze the temporal context of the representative features extracted from sub-epochs at both the intra- and inter-epoch levels. Their results showed that considering both the intra- and inter-epoch temporal context is effective to improve the performance of sleep scoring. Though the model achieved the state-of-the-art performance, the model used multi-channel signals and required data preprocessing via the short-time Fourier transform to extract the time-frequency domain data.
    
    In this paper, intra- and inter-epoch temporal context network (IITNet) is proposed for automatic sleep scoring on raw single-channel EEG. IITNet encodes each sub-epoch of multiple EEG epochs into the corresponding representative feature and analyzes their temporal context at both intra- and inter-epoch levels. IITNet is an end-to-end deep learning model based on one-step training without data preprocessing such as short-time Fourier transform or filterbank processing. A modified deep residual neural network (ResNet-50) \cite{he2016deep,he2016identity} is used to extract representative features from each epoch (half-minute EEG) at a sub-epoch level. Two layers of bidirectional long short-term memory (BiLSTM) \cite{schuster1997bidirectional} are used to learn the temporal context of the representative features in the forward and backward directions. The performance was investigated for three public datasets: SleepEDF, Montreal Archive of Sleep Studies (MASS), Sleep Heart Health Study (SHHS). Also, the influence of the number of input epochs, the sequence length (\(L\)), was investigated by increasing the sequence length from one to ten.

\section{IITNet: Intra- and Inter-epoch Temporal Context Network}
\label{sec:modelArch}
	
\subsection{Model Overview}

    IITNet is designed to classify the time-series data by extracting representative features at the sub-epoch level and analyzing their temporal context. In this study, the model is applied to sleep stage classification from raw single-channel EEG. When human sleep experts label each half-minute PSG (target epoch) with its corresponding sleep stage, they visually inspect the frequency characteristics and the sleep-related events such as spindles and K-complexes. Besides, they consider the sleep-related events in its neighboring epochs to check whether the relations of the events correspond to the transition rules \cite{tsinalis2016automatic}. Similarly, IITNet learns the sleep-related events by extracting representative features at the sub-epoch level and scores the sleep stages of the target epoch by capturing the contextual information from the sequence of the features.
        
    IITNet belongs to a convolutional recurrent neural network (CRNN) \cite{shi2017end} and consists of two main parts: CNN and RNN layers, as shown in Fig. \ref{fig:ModelArchitecture}. The CNN layers learn the representative features associated with the sleep-related events from the EEG. To train the deep CNN effectively, a modified ResNet-50 \cite{he2016deep} is used since its skip connections ensure that the higher layers can perform as good as the lower layer \cite{he2016identity}. For this reason, the ResNet has been widely used in deep networks as a promising feature extractor \cite{he2017mask,xie2018rethinking,lu2016hierarchical}.
    
\begin{figure}
    \centering
     \includegraphics[width=0.5\textwidth]{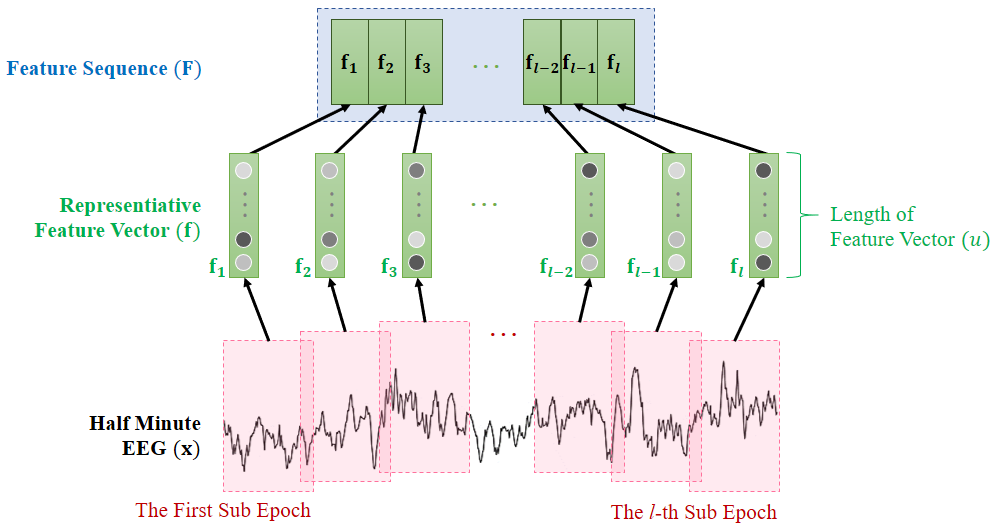}
     \caption{Conversion from an epoch signal into feature sequences in the CNN layers. Each sub epoch is encoded to its corresponding feature vector with the dimension of \(u=128\). Total \(l\) feature vectors are extracted and stacked into a feature sequence to be fed into the RNN layers.}
     \label{fig:receptive_field}
    \end{figure}

    In the RNN layers, two layers of bidirectional LSTM (BiLSTM) are employed to capture the temporal context from the representative features in both the forward and backward directions \cite{graves2013speech,hochreiter1997long,gers2002learning}. At the top of the model, a softmax classifier is placed to output the most appropriate sleep stage. Specifically, IITNet disassembles each half-minute epoch into \(l\) overlapped sub-epochs and encodes each sub-epoch as its corresponding representative feature, as shown in Fig. \ref{fig:receptive_field}. In the CNN layers, the epoch is converted to feature maps. Each column of the feature maps represents the sub-epoch feature corresponding to its receptive field. These \(l\) sub-epoch features are stacked from left to right in chronological order, and then a feature sequence is created for each epoch. The RNN layers analyze the temporal relation between the sub-epoch features.
        
    For IITNet, the input can be either single epoch or a series of successive epochs. Using only the target epoch as the input, the intra-epoch temporal context can be analyzed at the sub-epoch level. To consider both the intra- and inter-epoch temporal context, the sequence of the target epoch and its neighboring epochs should be fed at a time. In this study, the target epoch and its previous epochs are used as the input, which is practical for real-time sleep scoring in a smart bed or hospital bed because future epochs cannot be measured in advance. This is also clinically reasonable since human experts generally investigate the previous epochs to follow the AASM transition rules.

    \subsection{Intra-epoch Temporal Context Learning}
    \label{Intra-epoch Temporal Context Learning}

    For intra-epoch temporal context learning, IITNet takes a target epoch \(\mathbf{x}\) as an input to model the conditional probability \(p(y|\mathbf{x})\), where \(y\) is the true sleep stage.  In the CNN layers as shown in Fig. \ref{fig:ModelArchitecture}, IITNet extracts the representative features \(\mathbf{f}\) from the sub-epochs and produces a feature sequence \(\mathbf{F}\), which contains the sub-epoch features in chronological order as follows:
    \begin{equation} 
     \begin{split}
      \mathbf{F} = \{\mathbf{f}_{1}, \mathbf{f}_{2}, \cdots, \mathbf{f}_{l}\}, 
     \end{split}
    \end{equation}
    
    where \(\mathbf{f}_i\) is the representative feature vector of the \(i\)-th sub epoch, and \(l\) is the number of the sub epochs. The length of a feature vector is \(u\), which is the number of filters in the last CNN layer. Note that the learnable parameters of the CNN layers are shared. Through backpropagation in training, the parameters are updated with the average of the gradients computed for the sub-epochs. 
    
    In the RNN layers, the BiLSTM has hidden states of dimension \(u\) for each direction. The two BiLSTM layers process the feature vector with the previous hidden state in both the forward and backward directions, yielding the internal representation of the forward \(\overrightarrow{h_{t}}\) and backward \(\overleftarrow{h_{t}}\) contexts at time step \(t\), where \cite{graves2005framewise} 
    
    \begin{equation} \label{forward}
    \overrightarrow{\mathbf{h}_{t}} 
    = \overrightarrow{LSTM}(\mathbf{f}_{t}, \overrightarrow{\mathbf{h}_{t-1}}),
    \end{equation}
    \begin{equation} \label{backward}
    \overleftarrow{\mathbf{h}_{t}} 
    = \overleftarrow{LSTM}(\mathbf{f}_{t}, \overleftarrow{\mathbf{h}_{t+1}}).
    \end{equation}
    
    To predict \(y\), the last hidden states are concatenated to form the bidirectional context \(\overleftrightarrow{\mathbf{h}}\) of size \(2u\). This concatenated vector is fed into the fully connected layer to output \(p(y|\mathbf{x})\) for the target epoch. Finally, the softmax classifier labels the epoch with the most likely sleep stage.
    
    \subsection{Intra- and Inter-epoch Temporal Context Learning}
    \label{Intra- and Inter-epoch Temporal Context Learning}

    To consider the inter-epoch dependency at the sub-epoch level, IITNet takes the target epoch and its previous epochs as an input. The model scores the target epoch based on the temporal context in a series of feature sequences after encoding the target epoch and its previous \(L-1\) epochs at the sub-epoch level. Therefore, \(L\) is the number of epochs for the input. Formally, IITNet is trained to model the following conditional probability:
    \begin{equation}
    \begin{split}
        p(y_L|\mathbf{x}_{1}, \mathbf{x}_{2}, \mathbf{x}_3,  \cdots , \mathbf{x}_{L}),
    \end{split}
    \end{equation}
    
    where \(\mathbf{X}_L = \{\mathbf{x}_{1}, \mathbf{x}_{2}, \cdots , \mathbf{x}_{L}\} \) is a sequence of successive epochs, \(\mathbf{x}_L\) is the target epoch, \(\mathbf{x}_{1}, \mathbf{x}_{2}, \cdots , \mathbf{x}_{L-1}\) are the previous epochs, and \(y_L\) is the true sleep stage of the target epoch. Since it is practical to predict the latest sleep stage in the view of real-time sleep scoring, the the sleep stage the latest epoch is chosen as the true label.
    
    In the CNN layers, IITNet individually extracts the feature sequence for each epoch, as shown in Fig. \ref{fig:ModelArchitecture}. In other words, the CNN layers take the \(i\)-th epoch \(\mathbf{x}_i\) and produce a corresponding feature sequence \(\mathbf{F}_i\). At the top of the convolution layers, a series of feature sequences \(\mathbf{S}_L\) is created as follows:
    
    \begin{figure*}[ht!]
    \centering
      \includegraphics[width=\textwidth]{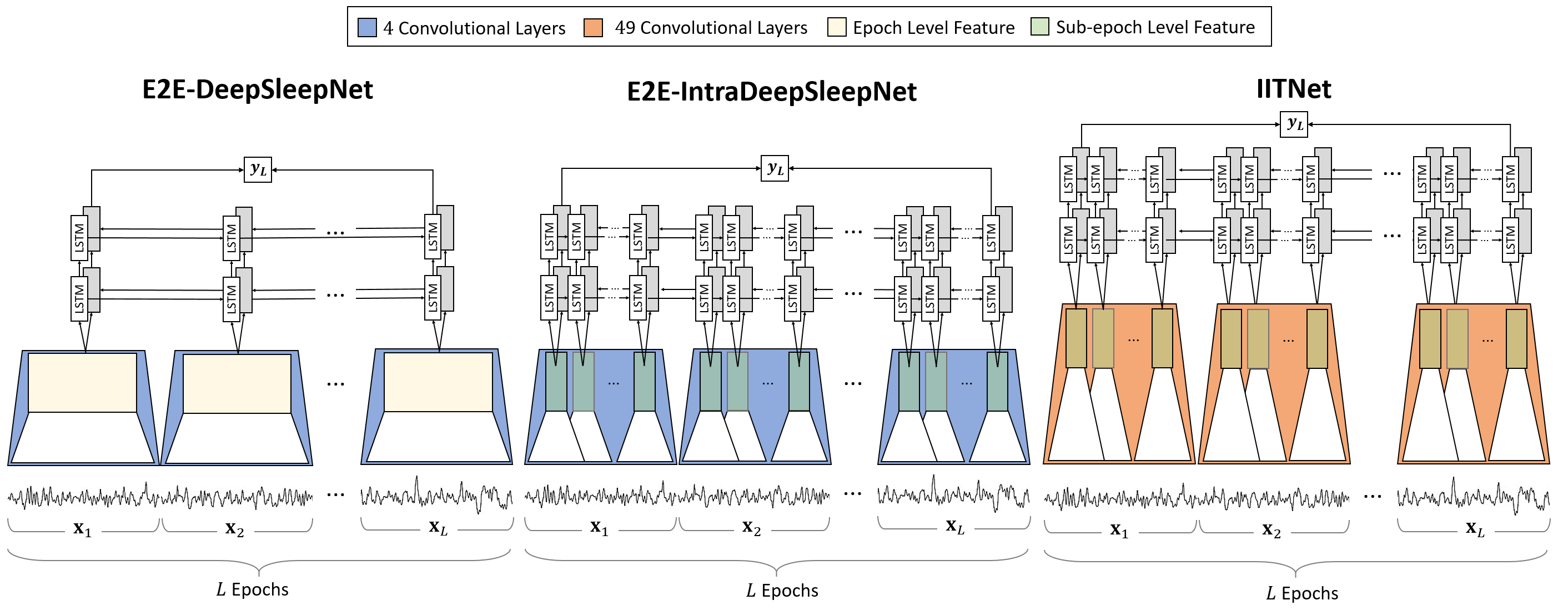}
      \caption{Model comparison between E2E-DeepSleepNet, E2E-IntraDeepSleepNet baseline and our proposed IITNet}  
      \label{fig:architecture_comparison}
    \end{figure*}
    
    \begin{equation} 
     \begin{split}
      \mathbf{S}_L & = \{\mathbf{F}_{1}, \mathbf{F}_{2}, \cdots, \mathbf{F}_{L}\}, \\
      &=\{\mathbf{f}_{1,1}, \mathbf{f}_{1,2}, \cdots, \mathbf{f}_{1,l},\\
      &\;\;\;\;\;\mathbf{f}_{2,1}, \mathbf{f}_{2,2}, \cdots,\mathbf{f}_{2,l}, \cdots, \\
      &\;\;\;\;\;\ \mathbf{f}_{L,1}, \mathbf{f}_{L,2}, \cdots, \mathbf{f}_{L,l}\}, 
     \end{split}
    \end{equation}
    where \(\mathbf{f}_{i,1}, \cdots, \mathbf{f}_{i,l}\) are the sub-epoch feature vectors corresponding to \(\mathbf{F}_i\). Accordingly, \(\mathbf{S}_L\) includes the entire representative features from the sub epochs in chronological order. Therefore, the RNN layers take the series of the feature sequence. The softmax classifier process \(\mathbf{S}_L\) in the same way as the intra-epoch temporal context learning.

	    \begin{table}[b]
    \centering
    \resizebox{\columnwidth}{!}{
    \footnotesize\addtolength{\tabcolsep}{-1pt}
    \begin{tabular}{@{}ccccccc@{}}
        \toprule
        Dataset & W & N1 & N2 & N3 & REM & Total \\
        \cmidrule(r){1-1} \cmidrule(r){2-6} \cmidrule{7-7} \\[-0.5em]
        SleepEDF & \begin{tabular}[c]{@{}c@{}} 8,285 \\ (20\%) \end{tabular} & \begin{tabular}[c]{@{}c@{}} 2,804 \\ (7\%) \end{tabular} & \begin{tabular}[c]{@{}c@{}} 17,799 \\ (42\%) \end{tabular} & \begin{tabular}[c]{@{}c@{}} 5,703 \\ (13\%) \end{tabular} & \begin{tabular}[c]{@{}c@{}} 7,717 \\ (18\%) \end{tabular} & 42,308 \\ [1em]

        MASS & \begin{tabular}[c]{@{}c@{}} 5,672 \\ (10\%) \end{tabular} & \begin{tabular}[c]{@{}c@{}} 4,524 \\ (8\%) \end{tabular} & \begin{tabular}[c]{@{}c@{}} 29,212 \\ (51\%) \end{tabular} & \begin{tabular}[c]{@{}c@{}} 7,567 \\ (13\%) \end{tabular} & \begin{tabular}[c]{@{}c@{}} 10,420 \\ (18\%) \end{tabular} & 57,395 \\ [1em]

        SHHS & \begin{tabular}[c]{@{}c@{}} 1,251,329 \\ (23\%) \end{tabular} & \begin{tabular}[c]{@{}c@{}} 217,508 \\ (4\%) \end{tabular} & \begin{tabular}[c]{@{}c@{}} 2,396,158 \\ (44\%) \end{tabular} & \begin{tabular}[c]{@{}c@{}} 739,212 \\ (14\%) \end{tabular} & \begin{tabular}[c]{@{}c@{}} 817,131 \\ (15\%) \end{tabular} & 5,421,338 \\ [1em]
        \bottomrule
    \end{tabular}}
        \caption{The class-wise number of epochs in the datasets}
        \label{table:dataset_info}
    \end{table}

	    \begin{table}[b]
    \centering
    \resizebox{\columnwidth}{!}{
    \footnotesize\addtolength{\tabcolsep}{-1pt}
    \begin{tabular}{@{}cccccccc@{}}
        \toprule
        \multirow{2}{*}{Dataset} & \multirow{2}{*}{Category}  & \multirow{2}{*}{$N_{s}$} & \multicolumn{4}{c}{Age} & \multirow{2}{*}{\begin{tabular}[c]{@{}c@{}} Epoch \\ per Subject \end{tabular}} \\
        \cmidrule(r){4-7}
                                 &                            &                          & Mean & Std & Min & Max  &    \\
        \cmidrule(r){1-1} \cmidrule(r){2-7} \cmidrule{8-8} \\[-0.5em]
                 & Men   & 10 & 28.3 & 2.2 & 26 & 32 &  \\
        SleepEDF & Women & 10 & 29.1 & 3.4 & 25 & 34 & 2,115 \\
                 & Total & 20 & 28.7 & 2.9 & 25 & 34 &  \\ [1em]
                 & Men   & 28 & 40.4 & 19.4 & 20 & 69 &  \\
        MASS     & Women & 34 & 44.2 & 18.6 & 20 & 69 & 926 \\
                 & Total & 62 & 42.5 & 18.9 & 20 & 69 &  \\[1em]
                 & Men   & 2,760 & 63.1 & 11.0 & 39 & 90 &  \\
        SHHS     & Women & 3,031 & 63.2 & 11.4 & 39 & 90 & 936 \\
                 & Total & 5,791 & 63.1 & 11.2 & 39 & 90 &  \\[1em]
        \bottomrule
    \end{tabular}}
        \caption{The demographic information of the datasets}
        \label{table:demographic_info}
    \end{table}

    \begin{table*}[ht!]
    \makegapedcells
    \footnotesize
    \begin{tabularx}{\linewidth}{c|c|Y|c|Y}
    \toprule
              & \multicolumn{2}{c|}{\bf{ResNet-50}}                        & \multicolumn{2}{c}{\bf{Modified ResNet-50 used in IITNet}}                \\ \hline
    Layer name                & blocks                            & output size            & blocks                & output size          \\ \hline
    conv1                     & 7 $\times$ 7, 64, stride 2                                 & 112 $\times$ 112       & 7 $\times$ 1, 64, stride 2       & 1500 $\times$ 1                 \\ \hline
    \multirow{2}{*}{\begin{tabular}[c]{@{}c@{}} \\ [0.2cm] conv2$\_$x\end{tabular}} & 3 $\times$ 3 max pool, stride 2  & \multirow{2}{*}{\begin{tabular}[c]{@{}c@{}} \\ [0.2cm] 56 $\times$ 56\end{tabular}} & 3 $\times$ 1 max pool, stride 2 & \multirow{2}{*}{\begin{tabular}[c]{@{}c@{}} \\ [0.2cm] 750 $\times$ 1\end{tabular}} \\ \cline{2-2} \cline{4-4}
                              & $\left[ \begin{tabular}[c]{@{}c@{}}1 $\times$ 1, 64\\ 3 $\times$ 3, 64\\ 1 $\times$ 1, 256\end{tabular} \right] $ $\times$ 3    &                & $\left[ \begin{tabular}[c]{@{}c@{}}1 $\times$ 1, 16\\ 3 $\times$ 1, 16\\ 1 $\times$ 1, 64\end{tabular} \right] $ $\times$ 3    &     \\ \hline
    conv3$\_$x                  & $\left[ \begin{tabular}[c]{@{}c@{}}1 $\times$ 1, 128\\ 3 $\times$ 3, 128\\ 1 $\times$ 1, 512\end{tabular} \right] $ $\times$ 4  & 28 $\times$ 28 & $\left[ \begin{tabular}[c]{@{}c@{}}1 $\times$ 1, 16\\ 3 $\times$ 1, 16\\ 1 $\times$ 1, 64\end{tabular} \right] $ $\times$ 4  & 375 $\times$ 1 \\ \hline
    \multirow{2}{*}{\begin{tabular}[c]{@{}c@{}} \\ [0.175cm] conv4$\_$x\end{tabular}} & \multirow{2}{*}{\begin{tabular}[c]{@{}c@{}} \\ [-0.175cm]$\left[\begin{tabular}[c]{@{}c@{}}1 $\times$ 1, 256\\ 3 $\times$ 3, 256\\ 1 $\times$ 1, 1024\end{tabular} \right] $ $\times$ 6 \end{tabular}}               & \multirow{2}{*}{\begin{tabular}[c]{@{}c@{}} \\ [0.2cm] 14 $\times$ 14\end{tabular}} & 3 $\times$ 1 max pool, stride 2 & \multirow{2}{*}{\begin{tabular}[c]{@{}c@{}} \\ [0.2cm] 94 $\times$ 1\end{tabular}}  \\ \cline{4-4} & & &  $\left[ \begin{tabular}[c]{@{}c@{}}1 $\times$ 1, 32\\ 3 $\times$ 1, 32\\ 1 $\times$ 1, 128\end{tabular} \right] $ $\times$ 6 & \\ \hline
    conv5$\_$x                  & $\left[ \begin{tabular}[c]{@{}c@{}}1 $\times$ 1, 512\\ 3 $\times$ 3, 512\\ 1 $\times$ 1, 2048\end{tabular} \right] $ $\times$ 3 & 7 $\times$ 7   & $\left[ \begin{tabular}[c]{@{}c@{}}1 $\times$ 1, 32\\ 3 $\times$ 1, 32\\ 1 $\times$ 1, 128\end{tabular} \right] $ $\times$ 3 & 47 $\times$ 1  \\ \hline
    out                       & average pool, 1000 fc, softmax                                                                                                 & 1 $\times$ 1   & dropout ($p=0.5$)                                                                                       & 47 $\times$ 1  \\ 
    \bottomrule
    \end{tabularx}
    \caption{Model comparison between ResNet-50 for ImageNet\cite{he2016deep} and the modified ResNet-50 used in IITNet. Down-sampling is performed by conv3$\_$1, conv4$\_$1, and conv5$\_$1 with a stride of 2. The left values in bracket represent the filter size and the right value indicates the number of filters. Output size is calculated on the assumption the that input for ResNet-50 is an image of 224$\times$224 and the input for IITNet is a 30-second EEG of 3000$\times$1 (SleepEDF)}
    \label{table:comp_resnet_ours}
    \end{table*}

\section{Experiments}
\label{sec:exp}
    \subsection{Datasets}
    To evaluate the sleep scoring performance of IITNet, three public datasets containing PSG records and their corresponding sleep stages labeled by human sleep experts were used: SleepEDF \cite{physiotoolkitphysionet,kemp2000analysis}, MASS \cite{o2014montreal}, and SHHS \cite{quan1997sleep}. Table \ref{table:dataset_info} lists the number of epochs in the datasets for the sleep stages and Table \ref{table:demographic_info} summarizes the demographic information of datasets, including gender distributions and age characteristics. Handcrafted feature extraction or signal processing was not employed for this study.

    \subsubsection{SleepEDF} 
    The SleepEDFx dataset (2013 version) contains two types of PSG record: SC for 20 healthy subjects without sleep-related disorders and ST for 22 subjects of a study on Temazepam effects on sleep. Each record includes two-channel EEGs from the Fpz-Cz and Pz-Oz channels, a single-channel EOG, and a single-channel EMG. Each half-minute epoch is labeled as one of eight classes (W, REM, N1, N2, N3, N4, MOVEMENT, UNKNOWN) according to R\&K rules. In this study, the single-channel EEGs (Fpz-Cz) in the SC (average subject age: \(28.7\) $\pm$ \(2.9\) years) were used since the Fpz-Cz has shown higher performance than Pz-Oz with deep-learning based approaches  \cite{tsinalis2016automatic, supratak2017deepsleepnet}. As the class W group was quite large compared to the others, only sixty epochs (thirty minutes) before and after the sleep period were used \cite{supratak2017deepsleepnet}. 
    
    \subsubsection{MASS} 
    The MASS dataset includes the PSG records of 200 subjects in five subsets: SS1, SS2, SS3, SS4, and SS5. These subsets are grouped according to the research and acquisition protocols. The dataset contains twenty-channel EEG, two-channel EOG, three-channel EMG, and single-channel ECG. Each epoch is labeled as one of five classes (W, REM, N1, N2, N3) according to AASM rules. In this study, F4-EOG (left) channel in the SS3 records (62 subjects, average subject age: \(42.5\) $\pm$ \(18.9\) years) was used.
    
    \subsubsection{SHHS}
    The SHHS dataset is a multi-center cohort study to investigate the effect of sleep-disordered breathing on cardiovascular diseases. The dataset consists of two rounds of PSG records: Visit 1 (SHHS-1) and Visit 2 (SHHS-2). Each record includes two-channel EEGs, two-channel EOGs, single-channel EMG, single-channel ECG, two-channel respiratory inductance plethysmography, position sensor data, light sensor data, pulse oximeter data, and airflow sensor data. In this study, the single-channel EEGs (C4-A1) in 5,791 records of the SHHS-1 were used. Each epoch is scored as either W, REM, N1, N2, N3, N4 using R\&K rule. More details are described in \cite{NSRR}.
    Note that some epochs in the datasets were labeled with MOVEMENT and UNKNOWN. They were excluded in this study because their prediction is beyond the scope of sleep stage classification. N3 and N4 stages were regarded as N3 stage according to AASM rules.
    
    \subsection{Baselines}
     Depending on the dataset processing methods, especially regarding whether using only in-bed or light-out parts of the dataset, a direct comparison between sleep scoring methods would not be a reasonably straightforward comparison; for example, including more segments with WAKE label can lead to high overall accuracy because the performance of sleep stage scoring methods on WAKE segments is usually better compared with the performance on other segments. On the other hand, a training method also affects performance. In order to fairly verify the effectiveness of the deep residual network \cite{he2016deep} and intra-epoch temporal context learning for sleep scoring, baseline experiments were conducted by modifying DeepSleepNet with an end-to-end (E2E) approach. First, E2E-DeepSleepNet was used to compare the performance in terms of model architecture, excluding the influence of a training method such as pre-training and fine-tuning. Secondly, E2E-IntraDeepSleepNet was used to confirm whether intra-epoch temporal context learning was effective in a shallow network. They were trained and evaluated on the same three datasets (SleepEDF, MASS, and SHHS) with the same pre-processing and training condition when sequence length (L) is 1, 4, 10, as shown in Fig. \ref{fig:architecture_comparison}. 

    \subsubsection{End-to-End DeepSleepNet (E2E-DeepSleepNet)}
    We used the DeepSleepNet as a deep learning baseline \cite{supratak2017deepsleepnet} and implemented it with the same architecture and parameter used in the paper. DeepSleepNet, which showed a fine performance in sleep stage scoring of the target epoch from a sequence of single-channel EEG signals, utilizes two parallel CNNs of small and large filters to extract time-invariant features and uses bidirectional LSTM to consider the sleep-stage transitions. For the comparison under the same condition, the model was trained in an end-to-end manner, similar to IITNet, instead of using a two-step training algorithm that finetunes the model using the sequential whole-night epochs after pre-training the CNN parts. This end-to-end DeepSleepNet is similar to the model experimented in \cite{phan2019seqsleepnet}.

    \subsubsection{End-to-End Intra-epoch Temporal Context DeepSleepNet (E2E-IntraDeepSleepNet)}
    To evaluate the effectiveness of intra-epoch temporal context learning, we modified the end-to-end DeepSleepNet baseline by introducing an intra-epoch temporal context learning, which was described in section \ref{Intra-epoch Temporal Context Learning} and \ref{Intra- and Inter-epoch Temporal Context Learning}. This end-to-end intra-epoch temporal context DeepSleepNet first extracts the sleep-related features from two CNN branches. The interpolation is performed after the CNN branches with larger filters to make the number of the sub-epochs of two CNN branches equal. Thereafter, features from two branches are concatenated, and two convolutional layers are applied channel-wise to half the length of the feature vector to form a feature sequence. In this way, sleep-related features can be analyzed in the sub-epoch level in RNN layers. Whereas the IITNet consists of ResNet-50 with residual connection and 49 convolutional layers, IntraDeepSleepNet-baseline uses a relatively shallow network with two parallel CNNs of four convolutional layers.

    \begin{figure*}[bh!]
    \centering
      \includegraphics[width=0.95\textwidth]{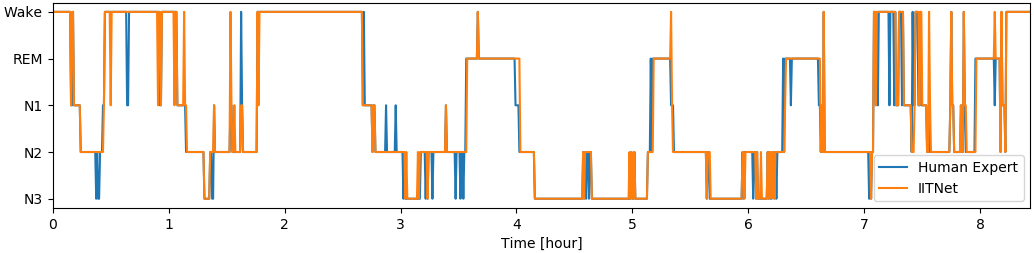}
      \caption{Hypnogram comparison between human expert and IITNet}  
      \label{fig:hypnogram}
    \end{figure*}    

    \subsection{Model Specifications}
    
    As described in Table \ref{table:comp_resnet_ours}, to handle one-dimensional time series EEGs, the one-dimensional operations of the modified ResNet-50 \cite{he2016deep} were used instead of the two-dimensional oper ations of convolution, max-pool, batch normalization. Furthermore, an additional max-pool was placed between the conv3$\_$x and conv4$\_$x layers to halve the feature sequence length. The global average pool layer was excluded, and a dropout layer (\(p=0.5\)) was added to the end of the CNN layers to prevent overfitting. In the RNN layers, two BiLSTM layers were adopted. The hidden state size of the BiLSTM in each direction was set to \(u=128\), which corresponds to the number of last convolutional filters.
    
    \subsection{Input for IITNet}
    The sequence length (\(L\)), the number of epochs as an input for IITNet, can be adjustable in IITNet. 
    To investigate the influence of the sequence length, training and evaluation were conducted by increasing the sequence length from one to ten in this study. The range was decided after considering the experimental result that there was no significant difference in the comparison of the sequence length of 10, 20, and 30 \cite{phan2019seqsleepnet}. Although some studies used multi channel input, this study only uses single channel EEG to consider versatile applications for time series data. Multi channel input is also possible via an ensemble model based on IITNet, which can show the sensitivity of each channel for each class. Since the channel characteristics is beyond the scope of this study, multi channel input is not dealt with in this study.
    
    \subsection{Model Training}
    Without any preprocessing, Adam \cite{kingma2014adam} optimizer was used with parameters $lr=0.005$, $\beta_1=0.9$, $\beta_2=0.999$, and $\epsilon=10^{-8}$. To avoid overfitting, L2-weight regularization was applied with \(wr=10^{-6}\). Any methods to balance out on data processing or model training were not used in this study although the state-of-the-art algorithms use balanced-class sampling \cite{supratak2017deepsleepnet, tsinalis2016automatic} or class-balanced loss \cite{vilamala2017deep} for mitigating class imbalanced-problems. In all the experiments, the batch sizes were 256, 128, 256 for SleepEDF, MASS, and SHHS, respectively. Early stopping was implemented by tracking the validation cost, such that the training was stopped when there was no validation cost improvement for ten successive training steps. For each cross-validation, the model that achieved the best validation accuracy was used for evaluation in the test set. The training process was implemented in Python 3.5.0 and PyTorch 0.4.0 \cite{paszke2017automatic}. On an RTX 2080 Ti, the training time of IITNet was, in total, approximately 10 h (SleepEDF) to 15 h (MASS), which was approximately 30 min for each fold. For the SHHS, the total training time was approximately 10 h. A single forward pass of IITNet took 11.7, 35.1, 79.4 ms when sequence length (\(L\)) is 1, 4, 10, respectively, which was calculated by averaging 2,000 trials.

    \subsection{Model Evaluation}
    For SleepEDF and MASS, \(k\)-fold cross-validation was conducted. When the number of subjects in a dataset was \(N_s\), the \(N_s/k\) records were used for model evaluation, and the other records were split into training and validation data. The test-set subjects were sequentially changed by repeating this process \(k\) times so that the evaluation was performed over all subjects. To be specific, 20-fold cross-validation was conducted for SleepEDF. In each fold, the numbers of the subjects for training, validation, and evaluation were 15, 4, and 1, respectively. For MASS, 31-fold cross-validation was performed. In each fold, the records of two randomly selected subjects that did not overlap with the other folds were used as the test set, and the remaining records were divided into training (45 subjects) and validation (15 subjects) sets. For SHHS, the subjects were randomly divided into training, validation, and test sets in the ratio 5:2:3 as performed in \cite{sors2018convolutional}.
    
    The IITNet performance was assessed according to the following criteria: per-class precision (PR), per-class recall (RE), per-class F1 score (F1), overall accuracy (Acc.), macro-averaging F1 score (MF1), and Cohen's kappa coefficient ($\kappa$) \cite{cohen1960coefficient}\cite{sokolova2009systematic}. In the case of a classification task, 
    
    \begin{equation}
        {\mathrm{PR}_{i} = \frac{\mathrm{e}_{ii}}{\sum_{j=1}^{N_{c}}{e_{ij}}},}
        \label{precision}
    \end{equation}
    
    \begin{equation}
        {\mathrm{RE}_{i} = \frac{\mathrm{e}_{ii}}{\sum_{j=1}^{N_{c}}{e_{ji}}},}
        \label{recall}
    \end{equation}
    where $e_{ij}$ is the element in the $i$-th row and $j$-th column of the confusion matrix and $N_c$ is the number of sleep stages (five in this study).
    
    \begin{equation}
    \mathrm{F1}_{i} = \frac{2}{\frac{1}{\mathrm{PR}_{i}} + \frac{1}{\mathrm{RE}_{i}}} = \frac{2\mathrm{PR}_{i}\mathrm{RE}_{i}}{\mathrm{PR}_{i} + \mathrm{RE}_{i}},
    \end{equation}
    
    \begin{equation}
    \mathrm{Overall Accuracy} = \frac{\sum_{i=1}^{N_{c}}\mathrm{e}_{ii}}{\sum_{i=1}^{N_{c}}\sum_{j=1}^{N_{c}}{e_{ij}}},
    \end{equation}
    
    \begin{equation}
    \kappa = \frac{p_{o}+p_{e}}{1-p_{e}} = 1 - \frac{1-p_{o}}{1-p_{e}}
    \end{equation}
    
    PR represents the precision with which the model distinguishes the sleep stage from the other stages. RE represents the accuracy with which the model predicts the sleep stage. Overall accuracy is the ratio of the correct predictions to the total predictions, which is an intuitive performance measure. Since F1 is calculated from the harmonic mean of the PE and RE, it can be more informative than overall accuracy, especially in the case of an imbalanced class distribution, i.e., sleep stages in PSG. The average of F1 corresponds to MF1 and $\kappa$ indicates the agreement between human expert (truth) and IITNet (prediction) in sleep scoring.

	\section{Results and Discussion}
	\label{sec:resultsanddiscussion}
    \subsection{Performance Comparison to State-of-the-Art Models}
    
    A hypnogram predicted by IITNet for one of the subjects is presented in Fig. \ref{fig:hypnogram}, in which the model predictions are in good agreement with human expert's scores. Specifically, Table \ref{table:SOTA} lists the performance of IITNet and the state-of-the-art models with the model architectures, approaches, input channel types, subject numbers, input types for the deep learning models, and the number of epochs simultaneously input for scoring the sleep stage of the target epoch. The baseline of IITNet is the case that the sequence length (\(L\)) is one. For all the datasets, IITNet achieved the performance comparable to the state-of-the-art models using the single-channel EEG although preprocessing was not used, the sequence length was shorter, and the target epoch and its previous epochs were only considered. The best overall accuracy, MF1, and $\kappa$ are $83.9\%$, $77.6\%$, $0.78$ for SleepEDF (\(L\)=10), $86.5\%$, $80.7\%$, $0.80$ for MASS (\(L\)=9), $86.7\%$, $79.8\%$, $0.81$ for SHHS (\(L\)=10), respectively. The differences of overall accuracy, MF1, and $\kappa$ between the best of IITNet and other state-of-the-art models are $+1.89\%p$, $+0.73\%p$, $+0.018$ for SleepEDF, $+0.28\%p$, $-1.05\%p$, $-0.004$ for MASS, $-0.06\%p$, $+1.32\%p$, $-0.003$ for SHHS, respectively. According to \cite{phan2019seqsleepnet}, SeqSleepNet-30 used the spectrogram of three channels (EEG, EOG, EMG) as the input and its overall accuracy, MF1, and $\kappa$ were $87.1\%$, $83.3\%$, $0.815$ for MASS. Although IITNet used the raw EEG signals instead of the spectrogram images, the performance was observed to be similar. The modified ResNet-50 could learn effectively the representative features related to the sleep events at the sub-epoch level. The features could be analyzed by the RNN via the two-layered BiLSTM at both the intra- and inter-epoch levels, which contributed to learning the transition rules human experts considered.
    
    Compared to the other state-of-the-art models, the main advantages of IITNet are efficiency and adjustability attributed to using single-channel raw EEG and controlling input length. The proposed model achieved comparable performance with state of the art by using raw single-channel EEG, although the other studies used multi-channel EEGs or spectrogram instead of the raw signal. Furthermore, the input length of the proposed algorithm can be adjustable according to diverse application purposes. This provides high feasibility compared to the other algorithm. Table \ref{table:SOTA} supports these strong points of the proposed algorithm. It is also feasible to score the sleep stages automatically in real-time via IITNet with a single-channel EEG sensor, which can be one of the essential technologies for healthcare 4.0, especially for the next generation treatment and diagnosis of sleep disorders. Furthermore, IITNet can be directly applied to classify various kinds of time-series data since the model is the end-to-end architecture without pre-training or preprocessing. On the other hand, it should be noted that the sampling frequency of the input cannot be changeable after the model is trained. In order to score sleep stages from the signal of the different sampling frequency, preprocessing such as down or upsampling is required. The compared models of other studies have the same limitation. To solve this, further study is necessary with the datasets of various sampling frequencies.
    \begin{table*}[ht!]
    \centering
    \footnotesize\addtolength{\tabcolsep}{-3.5pt}
    \begin{tabular}{@{}cccccccccccccccc@{}}
        \toprule
        \multicolumn{8}{c}{Method} & \multicolumn{3}{c}{Overall Metrics} & \multicolumn{5}{c}{Per-class F1 Score}\\ [-0.7em]
        \cmidrule(r){1-8} \cmidrule(r){9-11} \cmidrule{12-16}
        Dataset & Model & Architecture & Channel & Input & Approach & Sequence Length & Subjects  & Acc. & MF1 & $\kappa$ & W & N1 & N2 & N3 & REM \\
        \cmidrule(r){1-8} \cmidrule(r){9-11} \cmidrule{12-16} \\ [-0.5em]
        SleepEDF & \textbf{IITNet} & CNN + RNN & Fpz-Cz & Raw Signal & One-to-One & 1 & 20 & 80.6 & 72.1 & 0.73 & 84.7 & 29.8 & 86.3 & 87.1 & 72.8 \\ [0.7em]
        SleepEDF & \textbf{IITNet} & CNN + RNN & Fpz-Cz & Raw Signal & Many-to-One & 4 (3 past) & 20 & \textbf{83.6} & \textbf{76.5} & \textbf{0.77} & 87.1 & 39.2 & 87.7 & 87.7 & 80.9 \\ [0.7em]
        SleepEDF & \textbf{IITNet} & CNN + RNN & Fpz-Cz & Raw Signal & Many-to-One & 10 (9 past) & 20 & \underline{83.9} & \underline{77.6} & \underline{0.78} & 87.7 & 43.4 & 87.7 & 86.7 & 82.5 \\ [0.7em]
        SleepEDF & Supratak \cite{supratak2017deepsleepnet} & CNN + RNN & Fpz-Cz & Raw Signal & Many-to-One & \makecell{Whole previous \\ night epochs} & 20 & 82.0 & 76.9 & 0.76 & 84.7 & 46.6 & 89.8 & 84.8 & 82.4 \\ [0.3em]
        SleepEDF & Phan \cite{phan2018joint} & \begin{tabular}[c]{@{}c@{}} Multitask \\ 1-max CNN \end{tabular} & Fpz-Cz & Spectrogram & One-to-Many & 1 & 20 & 81.9 & 73.8 & 0.74 & - & - & - & - & - \\ [0.7em]
        SleepEDF & Vilamala \cite{vilamala2017deep} & CNN & Fpz-Cz & Spectrogram & Many-to-One & 5 (2 past, future) & 20 & 81.3 & 76.5 & 0.74 & 80.9 & 47.4 & 86.2 & 86.2 & 81.9 \\ [0.7em]
        SleepEDF & Tsinalis \cite{tsinalis2016automatic} & CNN & Fpz-Cz & Raw Signal & Many-to-One & 5 (2 past, future) & 20 & 74.8 & 69.8 & 0.65 & 43.7 & 80.6 & 84.9 & 74.5 & 65.4 \\ [0.4em]

        \cmidrule(r){1-8} \cmidrule(r){9-11} \cmidrule{12-16}
        MASS & \textbf{IITNet} & CNN + RNN & \begin{tabular}[c]{@{}c@{}} F4-EOG \\ (left) \end{tabular} & Raw Signal & One-to-One & 1 & 62 & 84.5 & 76.6 & 0.77 & 86.9 & 36.8 & 90.5 & 88.1 & 81.0 \\
        MASS & \textbf{IITNet} & CNN + RNN & \begin{tabular}[c]{@{}c@{}} F4-EOG \\ (left) \end{tabular} & Raw Signal & Many-to-One & 4 (3 past) & 62  & \textbf{86.2} & \textbf{80.0} & \textbf{0.79} & 85.2 & 51.8 & 91.4 & 86.9 & 84.5 \\
        MASS & \textbf{IITNet} & CNN + RNN & \begin{tabular}[c]{@{}c@{}} F4-EOG \\ (left) \end{tabular} & Raw Signal & Many-to-One & 10 (9 past) & 62 & \underline{86.3} & 80.5 & 0.79 & 85.4 & 54.1 & 91.3 & 86.8 & 84.8 \\

        MASS & Supratak \cite{supratak2017deepsleepnet} & CNN + RNN & \begin{tabular}[c]{@{}c@{}} F4-EOG \\ (left) \end{tabular} & Raw Signal & Many-to-One &\makecell {Whole previous \\ night epochs} & 62 & 86.2 & \underline{81.7} & \underline{0.80} & 87.3 & 59.8 & 90.3 & 81.5 & 89.3 \\
        MASS & Dong \cite{dong2018mixed} & Mixed RNN & \begin{tabular}[c]{@{}c@{}} F4-EOG \\ (left) \end{tabular} & \begin{tabular}[c]{@{}c@{}} Handcrafted \\ Features \end{tabular} & Many-to-One & 5 (4 past) & 62 & 85.9 & 80.5 & 0.79 & 84.6 & 56.3 & 90.7 & 84.8 & 86.1 \\
        MASS & Phan \cite{phan2018joint} & \begin{tabular}[c]{@{}c@{}} Multitask \\ 1-max CNN \end{tabular} & C4-A1 & Spectrogram & One-to-Many & 1 & 200 & 78.6 & 70.6 & 0.70 & - & - & - & - & - \\

        \cmidrule(r){1-8} \cmidrule(r){9-11} \cmidrule{12-16} \\ [-0.5em]
        SHHS & \textbf{IITNet} & CNN + RNN & C4-A1 & Raw Signal & One-to-One & 1 & 5,791 & 83.6 & 71.8 & 0.77 & 88.7 & 21.3 & 86.1 & 84.9 & 78.1 \\ [0.7em]
        SHHS & \textbf{IITNet} & CNN + RNN & C4-A1 & Raw Signal & Many-to-One & 4 (3 past) & 5,791 & \textbf{86.3} & \textbf{78.8} & \underline{\textbf{0.81}} & 90.0 & 45.2 & 88.2 & 84.9 & 85.9 \\ [0.7em]
        SHHS & \textbf{IITNet} & CNN + RNN & C4-A1 & Raw Signal & Many-to-One & 10 (9 past) & 5,791 & 86.7 & \underline{79.8} & 0.81 & 90.1 & 48.1 & 88.4 & 85.2 & 87.2 \\ [0.7em]
        SHHS & Sors \cite{sors2018convolutional} & CNN & C4-A1 & Raw Signal & Many-to-One & 4 (2 past, 1 future) & 5,728 & \underline{86.8} & 78.5 & 0.81 & 91.4 & 42.7 & 88.0 & 84.9 & 85.4 \\ [0.7em]
        \bottomrule
     \end{tabular}

     \caption{Performance comparison between IITNet and the state-of-the-art methods for automatic sleep scoring via deep learning. The underlined indicates the highest and the bold is the result of IITNet with the sequence length of 4.}
     \label{table:SOTA}
     \end{table*}

    \subsection{Influence of the Sequence Length (\(L\))}
    
\begin{figure*}[t]
    \centering
    \begin{subfigure}[b]{0.24\textwidth}
        \centering
        \includegraphics[width=\textwidth]{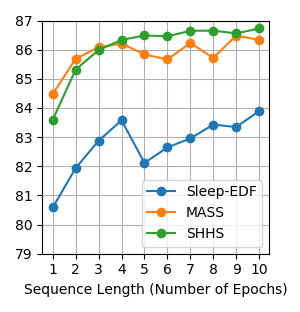}
        \caption[]%
        {{Overall Accuracy [\%]}}    
    \label{fig:criterion-acc}
    \end{subfigure}
    \begin{subfigure}[b]{0.24\textwidth}  
        \centering 
        \includegraphics[width=\textwidth]{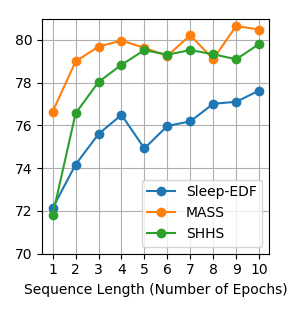}
        \caption[]%
        {{Macro-F1 [\%]}}    
    \label{fig:class-mf1}
    \end{subfigure}
    \begin{subfigure}[b]{0.24\textwidth}   
        \centering 
        \includegraphics[width=\textwidth]{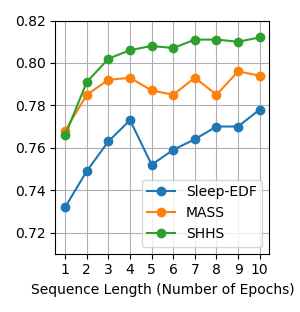}
        \caption[]%
        {{Cohen's Kappa ($\kappa$)}}    
    \label{fig:criterion-kappa}
    \end{subfigure}
    \begin{subfigure}[b]{0.24\textwidth}
        \centering
        \includegraphics[width=\textwidth]{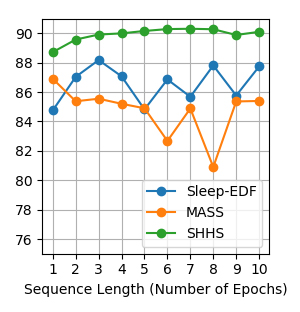}
        \caption[]%
        {{F1 of W [\%]}}    
    \label{fig:class-wake}
    \end{subfigure}
    
    \vskip\baselineskip
    \begin{subfigure}[b]{0.24\textwidth}
        \centering
        \includegraphics[width=\textwidth]{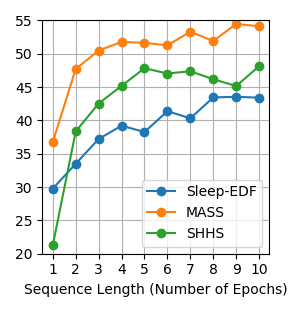}
        \caption[]%
        {{F1 of N1 [\%]}}    
    \label{fig:class-n1}
    \end{subfigure}
    \begin{subfigure}[b]{0.24\textwidth}  
        \centering 
        \includegraphics[width=\textwidth]{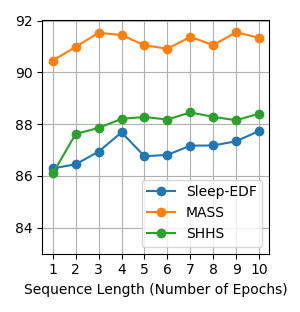}
        \caption[]%
        {{F1 of N2 [\%]}}    
    \label{fig:class-n2}
    \end{subfigure}
    \begin{subfigure}[b]{0.24\textwidth}   
        \centering 
        \includegraphics[width=\textwidth]{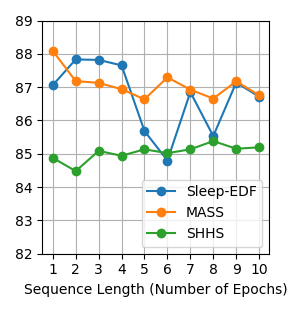}
        \caption[]%
        {{F1 of N3 [\%]}}    
    \label{fig:class-n3}
    \end{subfigure}
    \begin{subfigure}[b]{0.24\textwidth}   
        \centering 
        \includegraphics[width=\textwidth]{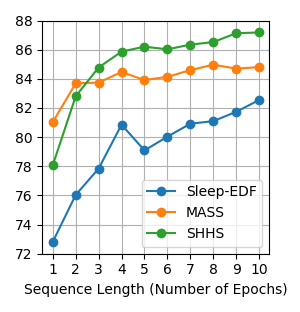}
        \caption[]%
        {{F1 of REM [\%]}}    
    \label{fig:class-rem}
    \end{subfigure}
    \caption[]
    {Performance of IITNet for SleepEDF, MASS, and SHHS according to the sequence length (\(L\))}
    \label{fig:performance}
    \end{figure*}

    The influence of sequence length (\(L\)) on the performance is shown in \ref{fig:performance}. For all the datasets, similar variation patterns can be seen. In Figs. \ref{fig:criterion-acc}-\ref{fig:criterion-kappa}, overall accuracy, MF1, and $\kappa$ consistently increase until \(L\) is four. After that, they oscillate according to \(L\). The smaller the fluctuation, the larger the dataset size (the number of subjects). The performance dropped in SleepEDF when the sequence length was 5 and 6. We think that the relatively small size of SleepEDF affected the drop since the models trained on a small dataset are more likely to result in high variance \cite{brain1999effect}. Moreover, the features of previous sleep stages before 2 minutes (more than 4 epochs) may be more difficult to be characterized. With insufficient training data, it is also hard for the model to learn the long ago features that affect the current sleep stage. The results show that this difficulty can be overcome with longer sequence length or larger datasets. Using four epochs (two minutes) as an input, the performance was still comparable to the state-of-the-art results for three datasets (SleepEDF, MASS, and SHHS) with overall accuracy (Acc.: $83.6\%$, $86.2\%$, $86.3\%$), macro F1-score (MF1: $76.5\%$, $80.0\%$, $78.8\%$), and Cohen's kappa ($\kappa$ : $0.77$, $0.79$, $0.81$).  The differences of overall accuracy, MF1, and $\kappa$ between IITNet (\(L\)) and the state-of-the-art models are $+1.58\%p$, $-0.41\%p$, $+0.013$ for SleepEDF, $+0.01\%p$, $-1.73\%p$, $-0.007$ for MASS, $-0.46\%p$, $+0.35\%p$, $-0.009$ for SHHS, respectively. When \(L\) increases from one to four, overall accuracy and MF1 improve by $2.99\%p$, $4.36\%p$ for SleepEDF, $1.73\%p$, $3.32\%p$ for MASS, and $2.74\%p$, $7.03\%p$ for SHHS, respectively. However, when \(L\) increases from four to ten, these two metrics increase by $0.32\%p$, $1.14\%p$ for SleepEDF, $0.13\%p$, $0.51\%p$ for MASS, and $0.39\%p$, $0.97\%p$ for SHHS, respectively. This results support that considering last two-minute epochs can be a reasonable choice to predict the sleep stage with efficiency and reliability. According to SeqSleepNet using the spectrogram of multi-channel PSG \cite{phan2019seqsleepnet}, the performance showed no significant difference when the sequence length was set to 10, 20, and 30. Other state-of-the-art models can be more compact, and their prediction can become faster by reducing the input length as recommended in this study.

    \begin{figure*}[!]
    \centering
        \includegraphics[width=\textwidth]{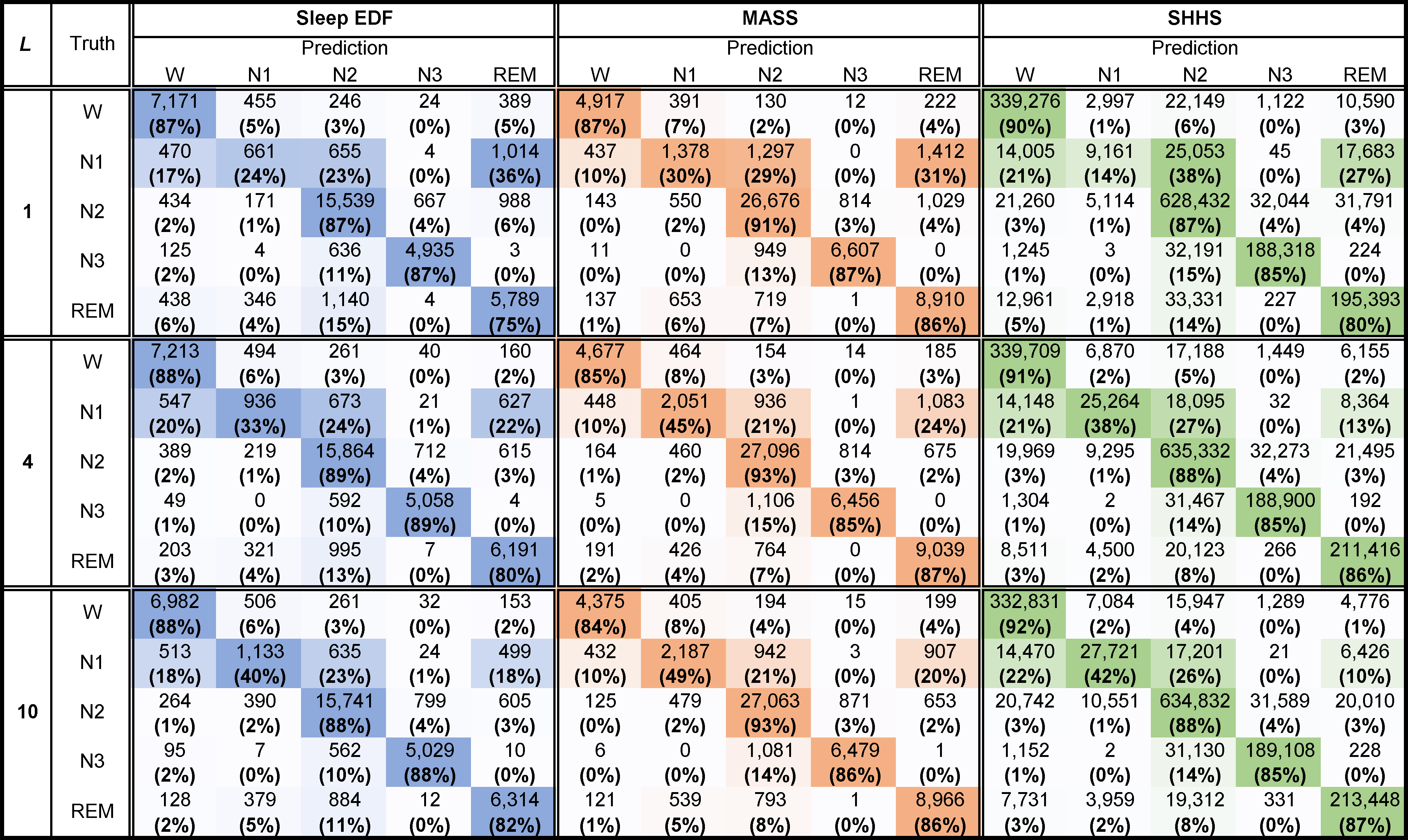}
        \caption{Confusion matrix of IITNet for SleepEDF, MASS, and SHHS with the sequence length (\(L\)) of one, four, and ten}  
        \label{fig:confusion}
    \end{figure*}

    \begin{table*}[]
        \centering
        \begin{tabular}{@{}c|ccc|ccc|ccc|ccc@{}}
        \toprule
        \multirow{2}{*}{Sequence Length} & \multirow{2}{*}{Model} & \multirow{2}{*}{Temporal Context} & \multirow{2}{*}{CNN} & \multicolumn{3}{c|}{SleepEDF} & \multicolumn{3}{c|}{MASS} & \multicolumn{3}{c}{SHHS} \\ \cmidrule(l){5-13} 
         &  &  &  & Acc. & MF1 & $\kappa$ & Acc. & MF1 & $\kappa$ & Acc. & MF1 & $\kappa$ \\ \midrule
        \multirow{3}{*}{1} & E2E-DeepSleepNet & Inter epoch & shallow & 80.9 & 73.1 & 0.74 & 84.4 & 76.9 & \textbf{0.77} & 80.8 & 68.2 & 0.72 \\
         & E2E-IntraDeepSleepNet & Intra + Inter epoch & shallow & \textbf{81.6} & \textbf{73.8} & \textbf{0.75} & \textbf{84.8} & \textbf{77.6} & \textbf{0.77} & 80.8 & 69.7 & 0.73 \\
         & IITNet & Intra + Inter epoch & deep & 80.6 & 72.1 & 0.73 & 84.5 & 76.6 & \textbf{0.77} & \textbf{83.6} & \textbf{71.8} & \textbf{0.77} \\ \midrule
        \multirow{3}{*}{4} & E2E-DeepSleepNet & Inter epoch & shallow & 82.0 & 76.2 & 0.75 & \textbf{86.8} & \textbf{81.8} & \textbf{0.80} & 83.9 & 76.0 & 0.77 \\
         & E2E-IntraDeepSleepNet & Intra + Inter epoch & shallow & 82.6 & 76.4 & 0.76 & 86.7 & 81.6 & \textbf{0.80} & 83.7 & 75.5 & 0.77 \\
         & IITNet & Intra + Inter epoch & deep & \textbf{83.6} & \textbf{76.5} & \textbf{0.77} & 86.2 & 80.0 & 0.79 & \textbf{86.3} & \textbf{78.8} & \textbf{0.81} \\ \midrule
        \multirow{3}{*}{10} & E2E-DeepSleepNet & Inter epoch & shallow & 82.7 & 77.4 & 0.76 & 86.5 & 81.1 & 0.80 & 85 & 78.2 & 0.79 \\
         & E2E-IntraDeepSleepNet & Intra + Inter epoch & shallow & 83.6 & \textbf{78.2} & \textbf{0.78} & \textbf{87.2} & \textbf{81.8} & \textbf{0.81} & 85.7 & 78.5 & 0.80 \\
         & IITNet & Intra + Inter epoch & deep & \textbf{83.9} & 77.6 & \textbf{0.78} & 86.3 & 80.5 & 0.79 & \textbf{86.7} & \textbf{79.8} & \textbf{0.81} \\ \bottomrule
        \end{tabular}
     \caption{Performance comparison between IITNet, E2E-DeepSleepNet baseline, E2E-IntraDeepSleepNet baseline for SleepEDF, MASS, and SHHS datasets. Bold indicates the highest results for the specific dataset and sequence length.}
     \label{table:baselines}
    \end{table*}
    
    \subsection{Per-class Performance Improvement}
    The F1 of N1, N2, and REM increases in a similar fashion to overall performance, whereas the enhancement in the F1 of W and N3 is not apparent. In Figs. \ref{fig:class-wake}-\ref{fig:class-rem}, F1 of N1 and REM steadily increases until \(L\) is four. F1 of N2 also increases until \(L\) is four except for MASS \(L=4\). After that, they oscillate according to \(L\). The confusion matrices for \(L\) of one, four, and ten are illustrated in Fig. \ref{fig:confusion}. On average for three datasets, in four epochs compared to single epoch, accuracy and MF1 increased by $2.47\%p$ and $4.93\%p$, respectively. Notably, MF1 of N1, N2, and REM increased by $16.1\%p$, $1.50\%p$, and $6.42\%p$, respectively. The results mean that the overall performance improvement was attributed to the enhanced prediction of N1, N2, and REM. The AASM recommends that sleep experts should consider the target and previous epochs, especially when labeling the sleep stage as N1 or REM. The results support that the transition rules were well trained in IITNet by the intra- and inter-epoch context learning at the sub-epoch level. On the other hand, no significant improvement in W and N3 indicates that they have less inter-epoch dependency than the other stages. According to the AASM, for the target epoch to be identified as W, N2, or N3, the sleep-related events or specific EEG signal activities of the target epoch should mainly be considered. 
    
    For further improvement in N1, adding a frequency-aware module on IITNet and increasing the sequence length can be included for the future work, as the mixed frequency in the range of 4-7.99 Hz is a characteristic of N1 \cite{berry2012aasm}. Modifying an IITNet into a sequence-to-sequence classification model with an multi-channel ensemble, as was accomplished in \cite{phan2018joint, phan2019seqsleepnet}, is also possible for a more elaborate classification. However, introducing an intra-epoch temporal context learning with the deep residual network for sleep scoring is our main focus in this article, and the suggested modification will be included in the future work.
   
    On the other hand, the results show that IITNet overcame the imbalanced nature of PSG datasets without any sampling method or learning technique. Typically, the PSG datasets have an imbalanced class distribution, e.g., the number of N1 is less than 10\% as shown in Table \ref{table:dataset_info}. To account for this, cost-sensitive learning or class-balanced sampling has been applied \cite{stephansen2017use,biswal2017sleepnet,dong2018mixed,supratak2017deepsleepnet,tsinalis2016automatic}. Although these kinds of methods improved the performance of certain classes; however, the overall performance became worse \cite{sors2018convolutional}. IITNet shows that using the sequence of less than ten epochs as the input enhanced both the overall and class-wise performance.

    \subsection{Performance Comparison to the baselines}
        The performance comparison between IITNet, E2E-DeepSleepNet and E2E-IntraDeepSleepNet when the sequence length \(L\) is 1, 4, and 10 for SleepEDF, MASS, and SHHS dataset is shown in Table \ref{table:baselines}. It is worth noting that all experiments with the baselines were conducted using the same dataset and under the same condition. In SleepEDF and MASS, introducing our proposed intra-epoch temporal context learning on E2E-DeepSleepNet tend to improve the sleep scoring performance considerably in both cases when the input is a single epoch and multiple-epoch. E2E-IntraDeepSleepNet performed better than E2E-DeepSleepNet when the input is a single epoch (\(L\)=1) with the margin of overall accuracy, MF1, and $\kappa$ between two models being $+0.7\%p$, $+0.7\%p$, $+0.01$ for SleepEDF, and $+0.6\%p$, $+0.2\%p$, $+0.01$ for MASS. Even for the multi-epoch input, the overall performance of E2E-IntraDeepSleepNet is better compared with that of E2E-DeepSleepNet, with the differences in overall accuracy being $+0.6\%p$ (\(L\)=4) and $+0.9\%p$ (\(L\)=10) for SleepEDF and $-0.1\%p$ (\(L\)=4) and  $+0.7\%p$ (\(L\)=10) for MASS, respectively. In SHHS, the overall accuracy of E2E-IntraDeepSleepNet was similar, i.e., $+0.0\%p$ (\(L\)=1) and $-0.2\%p$ (\(L\)=4) or better with $+0.7\%p$ (\(L\)=10), in comparison to E2E-DeepSleepNet. This verifies that considering intra-epoch temporal context by learning with sub-epoch level features can lead to a performance gain in sleep scoring and introduces a synergistic effect with the inter-epoch temporal context learning.
    
    In SHHS, exploiting a ResNet-50 for the representation learning was a key factor in the improvement of the sleep scoring performance. The overall accuracy of IITNet was significantly higher for all sequence lengths with a high margin $+2.8\%p, +2.8\%p$ (\(L\)=1), $+2.4\%p, +2.6\%p$ (\(L\)=4), $+1.7\%p, +1.0\%p$ (\(L\)=10) comparing with those for E2E-DeepSleepNet and E2E-IntraDeepSleepNet. This shows that a deeper neural network (49 convolutional layers) with a residual connection can yield a better sleep scoring performance than the shallow network of DeepSleepNet (4 convolutional layers) on a large scale dataset. However, for the SleepEDF and MASS, which are relatively small datasets compared with SHHS, employing a ResNet-50 did not always guarantee a performance improvement. Although IITNet in SleepEDF showed similar or higher overall accuracy compared with E2E-IntraDeepSleepNet with the margin of $-1.0\%p$ (\(L\)=1), $+1.0\%p$ (\(L\)=4), and $+0.3\%p$ (\(L\)=10), the overall accuracy of IITNet in MASS was lower than that of E2E-IntraDeepSleepNet’s with $-0.3\%p$ (\(L\)=1), $-0.5\%p$ (\(L\)=4), and $-0.9\%p$ (\(L\)=10). This suggests that using ResNet-50 as a feature extractor can enhance the sleep scoring performance when a sufficient number of epochs per subject are given (2,115 in SleepEDF), while a shallow network can be sufficient to learn sleep-related features when the number of epochs per subject is small (926 in MASS). Thus, the number of epochs and subjects should be considered when designing the depth of CNN in the sleep scoring network.

	\section{Conclusions}
	\label{sec:conclusions}
	
	Human sleep experts search the sleep-related events and consider the transition rules to score the sleep stage of an epoch. Motivated by this approach, a novel deep learning model named IITNet is proposed to score the sleep stage more accurately by considering the inter- and intra-epoch temporal contexts using raw single-channel EEG. The deep CNN based on a modified ResNet-50 extracts the sleep-related features and the RNN via two-layered BiLSTM learns the transition rules. Using ten epochs or less as an input, IITNet achieved the performance comparable to other state-of-the-art results for three public datasets: SleepEDF, MASS, and SHHS. The results show that the proposed temporal context learning at both the intra- and inter-epoch levels is effective to classify the time-series inputs. The overall performance was enhanced when the sequence length increased from one to four, which was attributed to the enhanced prediction of N1, N2, and REM.	However, the improvement was not significant above four epochs. Using the target epoch and its three previous epochs, the overall performance was still comparable to state-of-the-art results, which supports that considering last two-minute raw single-channel EEG can be a reasonable option to predict the sleep stage with efficiency and reliability. Other state-of-the-art models can be more compact, and their training can become faster by reducing the input length. Moreover, IITNet can be directly applied to predict or classify various kinds of time-series data for healthcare and well-being applications since the model is based on the end-to-end architecture without pre-training or preprocessing tailored to sleep scoring. 
	
	\section*{Acknowledgments}
    This work was supported by the Institute of Integrated Technology (IIT) Research Project through a grant provided by Gwangju Institute of Science and Technology (GIST) in 2019 (Project Code: GK11470).
	
	\ifCLASSOPTIONcaptionsoff
	\newpage
	\fi

	
	
	%
    \bibliography{references.bib}{}
    \bibliographystyle{IEEEtran}

\end{document}